\documentclass{article}
\usepackage{iclr2027_conference,times}

\usepackage{amsmath,amsfonts,bm}

\def\eqref#1{equation~\ref{#1}}
\def\1{\bm{1}}

\def\ve{{\bm{e}}}

\DeclareMathAlphabet{\mathsfit}{\encodingdefault}{\sfdefault}{m}{sl}
\SetMathAlphabet{\mathsfit}{bold}{\encodingdefault}{\sfdefault}{bx}{n}

\usepackage{hyperref}
\hypersetup{colorlinks=true, linkcolor=red, citecolor=blue, urlcolor=blue}
\hypersetup{pdftitle={LongCap: Hour-Scale Timestamped Audio Captioning via Chunk-wise Processing},
  pdfauthor={Jinchuan Tian, Sen Chen, Wenbin Huang, Chao Weng, Zhengwei Huang, Jiatong Shi, Yichen Lu, Haoran Wang, Shinji Watanabe, Cong Zhou}}
\usepackage{url}
\usepackage{amsmath}
\usepackage{amssymb}
\usepackage{booktabs}
\usepackage{siunitx}
\usepackage{multirow}
\usepackage{wrapfig}
\usepackage{placeins}
\usepackage{needspace}
\usepackage{xcolor}
\usepackage{graphicx}
\usepackage{alltt}
\usepackage{framed}
\newenvironment{promptbox}%
  {\par\addvspace{0.4em}\begin{framed}\ttfamily
  \scriptsize
  \raggedright
   \setlength{\parindent}{0pt}%
   \hyphenpenalty=10000 \exhyphenpenalty=10000 \emergencystretch=3em
   \noindent\ignorespaces}%
  {\par\end{framed}\addvspace{0.4em}}

\title{LongCap: Hour-Scale Timestamped Audio\\Captioning via Chunk-wise Processing}

\author{\textbf{Jinchuan Tian}$^{1*}$, \textbf{Sen Chen}$^{2*}$, \textbf{Wenbin Huang}$^{2*}$,
\textbf{Chao Weng}$^{2\dagger}$,\\
\textbf{Zhengwei Huang}$^{2}$, \textbf{Jiatong Shi}$^{2}$, \textbf{Yichen Lu}$^{2}$,\\
\textbf{Haoran Wang}$^{1}$,
\textbf{Shinji Watanabe}$^{1}$, \textbf{Cong Zhou}$^{2}$\\[0.4em]
$^{1}$Language Technologies Institute, Carnegie Mellon University\\
$^{2}$Independent Researcher\\[0.2em]
{\small $^{*}$Equal contribution\quad $^{\dagger}$Project lead}
}

\iclrfinalcopy
\begin{document}

\maketitle
\lhead{Preprint}

\begin{abstract}
Describing hours of audio requires accurate event timing and sustained
descriptive detail throughout the recording. We present
LongCap, a framework for timestamped audio captioning (TAC), with the
human-annotated LongCapBench and three diarization-inspired metrics:
Timeline Error Rate (TER), Granularity Error Rate (GER), and Caption
Similarity (tcpCapSim). LongCap interleaves audio chunks with structured
event captions, using relative timestamps and retained context to support
length extrapolation within a 1M-token inference context window.
Event-level reannotation restores caption detail,
while systematic studies examine model scaling, dependency supervision,
and self-distillation. LongCap-Diar achieves strong benchmark results
and extrapolates to 16 hours of concatenated audio despite training on
recordings no longer than 40 minutes. Our TAC models are competitive
with Qwen3.5-Omni-Plus on short clips, outperform evaluated baselines in
caption similarity and timeline accuracy in segment-level comparisons
on LongCapBench-TAC-Long, and caption full recordings approaching two hours. On LongCapBench-TAC,
9B and 35B-A3B offer favorable size--performance tradeoffs, while
122B-A10B substantially improves event-dependency recovery. These
findings support long-horizon general audio understanding and provide
a foundation for always-on voice agents.
\href{https://jctian98.github.io/longcap/}{Homepage}
\end{abstract}

\section{Introduction}
\label{sec:intro}

\begin{figure}[!htb]
\centering
\includegraphics[width=\linewidth]{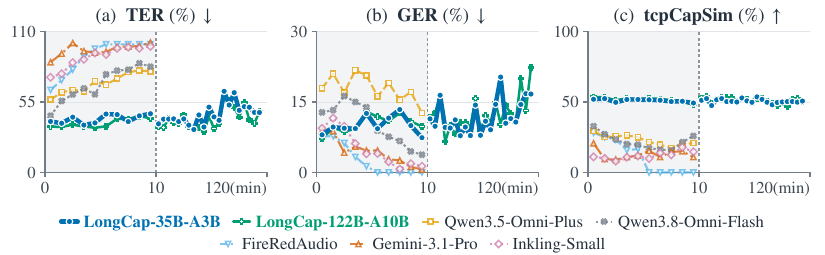}
\caption{Timestamped audio captioning (TAC) results on the LongCapBench-TAC-Long benchmark.
LongCap produces stable and comprehensive audio captions up to 2 hours, while competitors show severe degradation within 10 minutes.
(a) TER, (b) GER, and (c) tcpCapSim measure timeline
mismatch, granularity mismatch, and caption semantic similarity, respectively.}
\label{fig:teaser}
\vspace{-6pt}
\end{figure}

\begin{figure}[!t]
\centering
\vspace{-8pt}
\includegraphics[width=\linewidth]{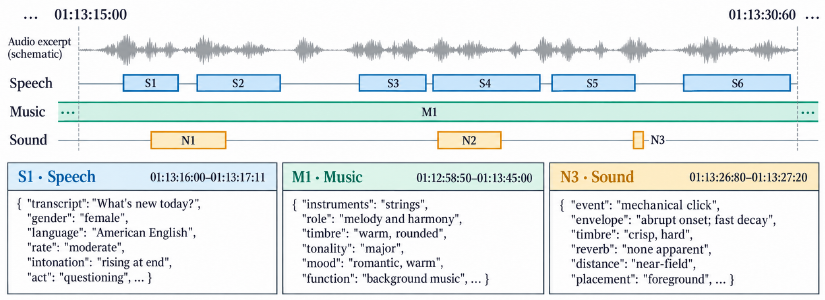}
\vspace{-8pt}
\caption{TAC format. A schematic excerpt of long-form audio illustrates
overlapping speech, music, and sound events with start/end timestamps and
structured attribute dictionaries.}
\label{fig:format}
\vspace{-6pt}
\end{figure}

Audio understanding is expanding from individual tasks toward
comprehensive descriptions of recordings. Speech recognition recovers
what was said \citep{radford2023whisper,kanda2020sot}, diarization identifies who spoke
when \citep{anguera2012diarizationreview,park2022diarizationreview}, and audio tagging detects sound
classes \citep{gemmeke2017audioset}. Audio captioning brings these aspects
together in a free-text description of a short clip
\citep{kim2019audiocaps,drossos2019clotho}. Recent captioners produce more
comprehensive descriptions of speech, music, and environmental sound,
using either free text or a structured format
\citep{qwen3omni2025,ma2026omnicaptioner,tian2026bagpiper}. A growing line of work goes
further by describing individual audio events together with their timestamps
\citep{kumar2026tac,rong2026audiomap,shao2026lataudio}. In this work, we
refer to this form as \emph{timestamped audio captioning} (TAC): the
output is a timeline of audio events, each with start and end timestamps
and a caption (Fig.~\ref{fig:format}). Joint speech recognition and diarization is a special case
in which each event caption contains only
the transcript and the speaker identity. To make the descriptions easier
to parse, we further require every caption to be
structured as key--value pairs over a closed, expert-designed key set, so
that each aspect of interest can be checked directly.

To extend this representation to hour-long recordings, we develop
LongCap together with an evaluation framework and a training recipe.
Inspired by diarization error rate (DER) and tcpWER
\citep{fiscus2007rt07,vonneumann2023meeteval} in speech recognition and diarization, we introduce Timeline Error Rate (TER)
to measure timeline mismatch, Granularity Error Rate (GER) to measure event
granularity mismatch, and Caption Similarity (tcpCapSim) to
measure semantic agreement between captions. We construct LongCapBench, a human-annotated benchmark
covering speech recognition, diarization, and TAC, with recordings
approaching two hours. Our modeling approach addresses two challenges:
\emph{holistic processing} \citep{qwen3omni2025,ghosh2026audioflamingonext}
increases the token distance between an event's audio and its description,
and models trained on short recordings can degrade on longer inputs
\citep{narayanan2019longform}. LongCap uses \emph{chunk-wise
processing} with chunk-relative timestamps, interleaving audio chunks
and event captions to keep emission local while retaining prior context.
We first develop and validate this design for speech recognition and
diarization, where supervision is abundant, then extend the recipe to
TAC. We examine model scaling from 0.8B to 122B parameters across pre-training
and post-training, assessing short-form accuracy and long-form
robustness. We further study \emph{density degradation}, explicit supervision
of long-range event dependencies, and self-distillation.

LongCap-Diar achieves the lowest macro-averaged recognition and
diarization errors among the evaluated systems
(Fig.~\ref{fig:diar_long}) and extrapolates to 16 hours of concatenated
audio within a 1M-token context window, despite training on recordings
no longer than 40 minutes
(App.~\ref{app:diar_results}). For TAC, LongCap is competitive with
Qwen3.5-Omni-Plus on short clips and captions full recordings approaching
two hours, surpassing evaluated competitors in caption similarity and
timeline accuracy in segment-level comparisons on LongCapBench-TAC-Long
(Fig.~\ref{fig:teaser}). Our scaling studies identify favorable
size--performance tradeoffs at 9B and 35B-A3B on LongCapBench-TAC
(Fig.~\ref{fig:sft_analysis}). The evaluated 122B-A10B model is
substantially stronger at recovering event dependencies, achieving
14.8\% long-form connectivity $F_1$ against silver references, compared
with 4.2\% for 35B-A3B
(Fig.~\ref{fig:dependency}). These results suggest different scaling
needs for event descriptions and dependencies, supporting long-horizon
general audio understanding and
always-on voice agents.

\section{Task Formulation: Timestamped Audio Captioning (TAC)}
\label{sec:task}

The formulation of \emph{timestamped audio captioning} (TAC) is inspired by the well-studied task of joint speech
recognition and diarization \citep{fiscus2007rt07,watanabe2020chime6,cornell2023chime7}, and extends its scope from speech transcription to general audio captioning.
Specifically, TAC localizes and describes every audio event in the recording. Given an
audio recording $X$, the model predicts a list of audio events
$E=[\ve_1,\dots,\ve_n]$, where each event $\ve_i=(t^{s}_i,t^{e}_i,c_i)$ is a
tuple of start timestamp, end timestamp, and a caption describing the sounds
within that span (Fig.~\ref{fig:format}).

\subsection{Formatting}
\label{ssec:formulation}
\label{ssec:attributes}

\textbf{Structured Caption}:
Unlike prior work that writes the caption as free-running natural language
\citep{qwen3omni2025,ma2026omnicaptioner}, we require every event caption to
be \emph{structured}: each caption is explicitly a set of key--value pairs, in
which the keys come from a pre-defined closed set and each value is a short
piece of precise natural language. The structure makes the
description explicit and easy to parse: each key of interest can be accessed and analyzed directly.

\textbf{Attribute Key Set}:
To define the attribute key set, we collaborated with audio professionals from the
speech, music, and film industries. The experts collaborated to make the keys
cover common aspects of audio information while avoiding
being duplicated, ambiguous or over-professional, converging to 62 keys.
A key appears only when it is applicable and is otherwise absent.
All attributes, together with the event timestamps, are assembled into a nested
JSON object and serialized as plain text, so the TAC task remains an
audio-to-text conversion task: it models the conditional distribution
$P(E\mid X)$. The detailed design of these keys is given in
App.~\ref{app:taxonomy}.

\subsection{Scoring}
\label{ssec:metrics}

\begin{figure}[t]
    \centering
    \includegraphics[width=\linewidth]{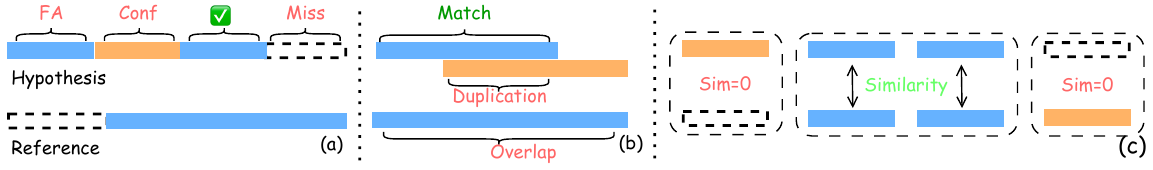}
    \caption{TAC scoring metrics.
    (a) TER measures timeline mismatch, (b) GER measures event granularity
    mismatch, and (c) tcpCapSim measures time-constrained caption similarity.}
    \label{fig:overview}

\end{figure}

Inspired by the Diarization Error Rate (DER)
\citep{fiscus2007rt07} and tcpWER
\citep{vonneumann2023meeteval} in speech recognition and diarization, we derive the \emph{Timeline Error Rate}
(TER) and \emph{tcpCapSim} for TAC, and additionally introduce
\emph{Granularity Error Rate} (GER) for event granularity mismatch.
Figure~\ref{fig:overview} illustrates TER, GER, and tcpCapSim in panels
(a)--(c), respectively; more detailed definitions are given
in App.~\ref{app:metric_formula}. All three compare predictions
$\hat{E}=[\hat{\ve}_1,\dots,\hat{\ve}_{\hat{n}}]$ with reference events
$E=[\ve_1,\dots,\ve_n]$.

\textbf{Event Grouping}:
DER and tcpWER optimize over speaker permutations, with utterances already
grouped by speaker labels. General audio events lack equivalent identity
labels, so their correspondence must be inferred from semantic content.
To approximate enumeration of possible groupings, a text LLM proposes
$M{=}3$ candidates $\mathcal{G}^{(m)}$. Each event belongs to exactly one
group; groups may contain multiple events from either side or have one
side empty. Each metric is evaluated under all three groupings and
independently retains its best score.

\textbf{TER~($\downarrow$)}: TER measures timeline mismatches for general audio events; its
name avoids confusion with speaker diarization. We accumulate miss duration
$T_\mathrm{miss}$ for too few active predicted groups, false alarm duration
$T_\mathrm{fa}$ for too many, and confusion duration $T_\mathrm{conf}$ for
remaining simultaneous activity in different groups. We normalize their sum
by $T_\mathrm{ref}$, the reference activity duration summed across groups
after merging overlapping intervals within each group.

\textbf{GER~($\downarrow$)}: GER measures splitting, merging, and repetition of events
through one-to-one matching: each event can match only one counterpart.
A split or merge can therefore preserve time coverage while reducing
matched overlap. Summed over groups, $T_\mathrm{overlap}$ is the overlap
of reference and prediction activity unions; $T_\mathrm{match}$ is the
maximum one-to-one matched overlap; and $T_\mathrm{dup}$ is summed
prediction event duration minus its union duration, accounting for
overlapping predictions.
\begin{equation}
\mathrm{TER}=\min_{\mathcal{G}^{(m)}}\;
\frac{T^{(m)}_\mathrm{miss}+T^{(m)}_\mathrm{fa}+T^{(m)}_\mathrm{conf}}{T^{(m)}_\mathrm{ref}},
\qquad
\mathrm{GER}=\min_{\mathcal{G}^{(m)}}\;
\frac{T^{(m)}_\mathrm{overlap}-T^{(m)}_\mathrm{match}+T^{(m)}_\mathrm{dup}}{\sum_{i=1}^{n}\big(t^{e}_i-t^{s}_i\big)}.
\label{eq:ter}
\end{equation}

\textbf{tcpCapSim~($\uparrow$)}: A group receives similarity credit only if a reference
event and a predicted event overlap or are at most five seconds apart.
We pool values by scored attribute key on each side and ask a text LLM to
compare each shared key's values for semantic agreement, scoring them in
$[0,1]$. A missing counterpart or failed time
constraint gives zero credit. Precision $P^{(m)}$ and recall $R^{(m)}$
divide the summed attribute scores by predicted and reference attribute
counts, respectively, across all groups, counting each key once per group
on each side. We take their arithmetic mean, yielding
$\mathrm{tcpCapSim}=\max_{\mathcal{G}^{(m)}}(P^{(m)}+R^{(m)})/2$.

\subsection{LongCapBench}
\label{ssec:benchmark}

We curate \emph{LongCapBench} to evaluate speech recognition and diarization
alongside TAC, with short- and long-form subsets for both settings.
Table~\ref{tab:longcapbench} summarizes the benchmark statistics.
All labels are finalized by human annotators.
Benchmark composition is described in App.~\ref{app:benchmark}.

TAC annotation follows a two-stage workflow. Starting from Qwen3.5-Omni-Plus
\citep{qwen35omni2026} predictions, annotators first correct event timestamps,
then verify each event's attributes. This process presents two main challenges.
First, unaided listeners often miss less salient events; model-generated
proposals guide event discovery and make verification more reliable and
efficient. Second, the breadth of audio attributes exceeds the expertise
a generalist can acquire through brief training. We therefore route data
batches through specialist groups, each reviewing a small set of related
attributes. Review requiring specialist expertise, particularly in music,
becomes a throughput bottleneck.
Comparing drafts with returned annotations, approximately 65\% and 53\%
of matched start/end boundaries differ on TAC-short and TAC-long,
respectively; 58\% and 91\% of attribute entries are added, removed, or
modified before final benchmark curation. The benchmark therefore uses
human-reviewed and corrected references, and we retain Qwen3.5-Omni-Plus
as an evaluation baseline despite its role in generating the initial drafts.
App.~\ref{app:benchmark} describes the annotation workflow.

\begin{table}[t]
\caption{Statistics of LongCapBench.
Recordings are evaluation clips.}
\label{tab:longcapbench}
\centering
\small
\setlength{\tabcolsep}{4pt}
\sisetup{detect-all, group-separator={,}, group-minimum-digits=4,
    table-number-alignment=right}
\scalebox{0.8}{%
\begin{tabular*}{\linewidth}{@{\extracolsep{\fill}}l
    S[table-format=4.0] S[table-format=2.2]
    S[table-format=3.2] S[table-format=3.2] S[table-format=5.0]@{}}
\toprule
Subset & {Recordings} & {Volume (h)} & {Mean length (min)} & {Maximum length (min)} &
{Utterances/events} \\
\midrule
Diar.-short & 442 & 53.01 & 7.20 & 11.10 & 44860 \\
Diar.-long & 15 & 15.02 & 60.06 & 78.16 & 4068 \\
TAC-short & 1500 & 18.50 & 0.74 & 1.00 & 9380 \\
TAC-long & 4 & 7.05 & 105.70 & 117.05 & 6520 \\
\bottomrule
\end{tabular*}
}
\end{table}

\section{LongCap}
\label{sec:method}

\textbf{Architecture}:
LongCap instantiates a TAC model with standard components: Qwen3.5
backbone language models (0.8B, 2B, 9B, 35B-A3B, and 122B-A10B), the
Qwen3-Omni-Captioner audio encoder \citep{qwen3omni2025}
feeding the backbone through a linear adaptor, and a cross-entropy loss
on the text positions.
All parameters are unfrozen during training. Qwen3.5 uses hybrid linear
and full attention and natively supports a 256k-token context
\citep{qwen352026}.
For contexts beyond 256k tokens, we use YaRN \citep{peng2023yarn} to extend
the context window to 1M tokens.

\Needspace{14\baselineskip}
\subsection{Chunk-wise Processing}
\label{ssec:chunkwise}

\begin{wrapfigure}{r}{0.30\linewidth}
\centering
\vspace{-12pt}
\includegraphics[width=\linewidth]{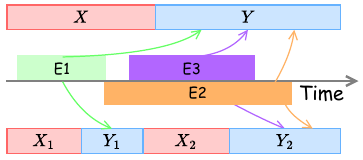}
\caption{Holistic (top) versus chunk-wise (bottom) processing of a long
recording.}
\label{fig:pipeline}
\end{wrapfigure}
Since the output $E$ is serialized as plain text (\S\ref{ssec:attributes}),
modeling $P(E\mid X)$ reduces to audio-conditioned language
modeling. The prevailing
recipe \citep{qwen3omni2025,ghosh2026audioflamingonext} is \emph{holistic
processing}: the whole recording $X$ is encoded into the context first, and
only then is $E$ decoded autoregressively, conditioned on the entire
recording.
LongCap instead adopts \emph{chunk-wise processing}, which interleaves
perception and emission (Fig.~\ref{fig:pipeline}). The recording is split into consecutive,
non-overlapping chunks $X=[X_1,\dots,X_K]$. Each chunk
$X_k$ starts at $\tau^{s}_k$ and ends at $\tau^{e}_k$; its label $Y_k$
contains all audio events that end within the chunk.\footnote{Events spanning
multiple chunks are assigned only to the chunk in which they end.} The
model then reads and writes the interleaved sequence
$[X_1,Y_1,X_2,Y_2,\dots,X_K,Y_K]$, retaining earlier audio chunks and generated
labels in context:\footnote{Conditioning each $Y_k$ on
$X_{\le k}$ rather than the full $X$ is the causal assumption of chunk-wise
processing.}
\begingroup
\begin{equation}
P(E\mid X)=\prod_{k=1}^{K}P\big(Y_k\mid X_{\le k},\,Y_{<k}\big).
\label{eq:chunkwise}
\end{equation}
\endgroup

\textbf{Comparison to Holistic Processing}:
Compared with holistic processing, chunk-wise processing offers two advantages
on hour-long recordings.
\emph{Local caption emission}: holistic processing emits an event's
description only after the entire recording and all earlier output tokens,
turning every prediction into long-context retrieval, where accuracy is known
to degrade \citep{liu2024lostmiddle,hsieh2024ruler}. Chunk-wise processing
emits each description after the chunk containing the event's end,
following chunk-local speech decoding \citep{jia2024speechllmxl}.
This keeps emission near the event endpoint while retaining earlier
chunks for long-range evidence.
\emph{Reduced length bias}: generation can degrade beyond training lengths
in language modeling \citep{press2022alibi} and speech recognition
\citep{narayanan2019longform}. Chunk-wise processing repeats the same local emission step on longer
recordings, supporting length extrapolation \citep{radford2023whisper,jia2024speechllmxl}.
This ability relies on the base text model's long-context capacity, which we
preserve during training (\S\ref{ssec:pretrain}).

\subsection{Long-Form Speech Recognition and Diarization}
\label{ssec:pretrain}

Speech recognition and diarization is a reduced version of TAC: each
event caption keeps only two keys, the speaker identity and the transcript
(\S\ref{ssec:formulation}). Its
supervision, however, is far more abundant than that of the general TAC
task, which makes it an ideal test bed for ablating chunk-wise processing in the long-form setup. Our ablations on this task, presented in
\S\ref{ssec:exp_diar}, justify three design choices:

\textbf{Relative timestamps}:
Absolute timestamps become unreliable beyond the training duration range
in our experiments (\S\ref{ssec:exp_diar}). Following Whisper's
window-relative timing \citep{radford2023whisper}, we express event
timestamps $t^{s}_i,t^{e}_i$ relative to the start of their chunk.\footnote{Recording-level event timestamps are recovered by adding
the chunk start to the relative predictions. An event crossing the chunk's
left boundary carries a negative start.}

\textbf{Chunk size}:
We compare a fixed chunk size of 32\,s with random sampling from a continuous
uniform distribution $\mathcal{U}(8,32)$\,s or the discrete set
$\{8,16,24,32\}$\,s. The chunking strategy has only a marginal impact on
performance. We adopt the discrete set $\{8,16,24,32\}$\,s, randomly sampling each chunk's duration
during both training and inference.

\textbf{Long-form text}:
To retain the base model's long-context and text capabilities, we include
long-form text in the training mixture, following text-mixing strategies
in long-context and speech-LLM training
\citep{gao2024prolong,peng2024voicetextblender}.

Our diarization model is trained at most 20 minutes, but can extrapolate to 16 hours (concatenated LongCapBench-Diar-Long examples) with stable WER, although the accumulated speaker count is far beyond the training distribution.

\subsection{Migration to Timestamped Audio Captioning}
\label{ssec:sft}

We initialize each LongCap model directly from a pretrained Qwen3.5 text
backbone and the Qwen3-Omni-Captioner audio encoder. Before the optional
extensions below, all training recordings are at most 20 minutes long.

\subsubsection{Pre-Training}
TAC requires two complementary capabilities: timestamp prediction and
audio attribute prediction. Because comprehensive TAC supervision is
scarce, we develop these capabilities through multi-task pre-training.
Speech recognition and diarization on recordings of up to 20 minutes
provide timestamp supervision, while natural-language captions paired
with audio chunks of up to 30 seconds teach audio attribute prediction.
These annotated chunks are spliced into training recordings of up to
10 minutes in total.
Following \S\ref{ssec:pretrain}, we also include long-form text sequences
of up to 256k tokens to preserve long-context capabilities. We mix
diarization, captioning, and text data in a 1:1:1 ratio by token count.
The resulting model is named \texttt{LongCap-Base}.
Data curation is described in App.~\ref{app:data}.

\subsubsection{Supervised Fine-Tuning (SFT)}
We refine the model and its training data in two stages, followed by two
optional extensions. We conduct the refinement with the 35B-A3B model and
use the resulting data to train models at all backbone sizes. During SFT,
captioning, diarization, and text data are mixed in a $3:1:1$ ratio by token count.
All SFT data volumes and mean durations below refer to audio retained
after filtering.

\textbf{Seed supervision}:
We curate 15k hours of audio, with recordings averaging 49 seconds and
lasting at most 10 minutes. We split each recording into clips of at most
one minute and query
Qwen3.5-Omni-Plus to generate structured captions in the format of
\S\ref{ssec:formulation}. The clips are annotated independently,
and their annotations are reassembled into recording-level labels.
Fine-tuning \texttt{LongCap-Base} on these labels yields
\texttt{LongCap-SFT-1}, establishing basic TAC capabilities.

\textbf{Alleviating density degradation}:
Both Qwen3.5-Omni-Plus and \texttt{LongCap-SFT-1} exhibit
\emph{density degradation}: later events within a chunk contain fewer
filled attributes (Fig.~\ref{fig:sft_analysis}c).
To restore event-level detail, we curate another 10k hours of audio, with
recordings averaging 49 seconds and lasting at most 10 minutes, and annotate them with
\texttt{LongCap-SFT-1}. Using the predicted timestamps, we extract each
event's audio span and query Qwen3.5-Omni-Plus to annotate it independently.
Fine-tuning \texttt{LongCap-SFT-1} on these refined annotations
yields \texttt{LongCap-SFT-2}, which maintains more consistent attribute
density across event positions.

\textbf{Optional 1: Length Expansion}:
The initial training recipe uses recordings no longer than 20 minutes.
To test whether longer training examples improve extrapolation to two-hour
recordings, we curate an additional 20k hours of audio covering diverse
acoustic content. Approximately 10k hours come from recordings longer than
20 minutes, with a maximum duration of 40 minutes. We use
\texttt{LongCap-SFT-2} to annotate this corpus; the longest resulting
audio--caption sequences approach the 256k-token training context limit.

\textbf{Optional 2: Event Dependency Modeling}:
Although the model retains the full audio and caption history, its labels
largely originate from short-window annotation. To make cross-event
dependencies explicit, we use GLM-5.3-Flash \citep{zai2026glm53flash} to analyze the event sequence
and, where appropriate, assign each event a semantically related earlier
event as its \emph{ancestor}. We include both the current event's
ID and its labeled ancestor in the JSON string used as the training target.
Following these links recursively forms event chains that describe how
events evolve over time, providing explicit supervision for long-range
dependencies.

\section{Experiments}
\label{sec:exp}

\subsection{Experimental Setup}
\label{ssec:setup}

\textbf{Training and Inference}:
We train all five backbone sizes following \S\ref{ssec:sft}, with
100k continued pre-training steps and two SFT epochs at a constant
learning rate of $1\times10^{-5}$. Context parallelism supports
256k-token training sequences. Inference uses a customized vLLM
implementation \citep{kwon2023vllm} with greedy decoding.
Hardware, hyperparameters, and decoding safeguards are detailed in
App.~\ref{app:training_inference}.

\textbf{Benchmarks}:
We evaluate speech recognition and diarization on AliMeeting
\citep{yu2022m2met}, AISHELL-4 \citep{fu2021aishell4}, AMI-IHM and AMI-SDM,
four CHiME-6 sessions
\citep{watanabe2020chime6}, and the LongCapBench-Diar subsets
(\S\ref{ssec:benchmark}). TAC is evaluated on the LongCapBench-TAC subsets
(\S\ref{ssec:benchmark}).
For general audio understanding, we use the MMAU test-mini subset
\citep{sakshi2024mmau} and MMAR \citep{ma2025mmar}.
Following Omni-Captioner \citep{ma2026omnicaptioner}, we first caption each
recording with the TAC model, then use a text LLM to answer the benchmark
questions from the caption alone.

\textbf{Competitors}:
For speech recognition and diarization, our baselines are MOSS Transcribe Diarize
\citep{moss2026transcribe}, VibeVoice-ASR \citep{peng2026vibevoiceasr},
Doubao ASR 2.0 \citep{doubaoasr2026}, and Qwen3.8-Omni-Flash
\citep{qwen38omni2026}.
For TAC, we compare against Qwen3.5-Omni-Plus \citep{qwen35omni2026},
Qwen3.8-Omni-Flash \citep{qwen38omni2026}, Gemini-3.1-Pro
\citep{gemini31pro}, FireRedAudio \citep{shen2026fireredaudio}, and
Inkling-Small \citep{inkling2026}, using commercial APIs or released checkpoints.
Differences in output format complicate direct comparisons with LAT
\citep{shao2026lataudio} and AF-Next \citep{ghosh2026audioflamingonext};
we compare their native caption content in App.~\ref{app:native_caption_results}.
App.~\ref{app:setup} describes LongCap data, training, and inference;
App.~\ref{app:metric_formula} specifies scoring, and
App.~\ref{app:additional_results} provides additional comparison protocols.

\begin{figure}[!t]
\centering
\includegraphics{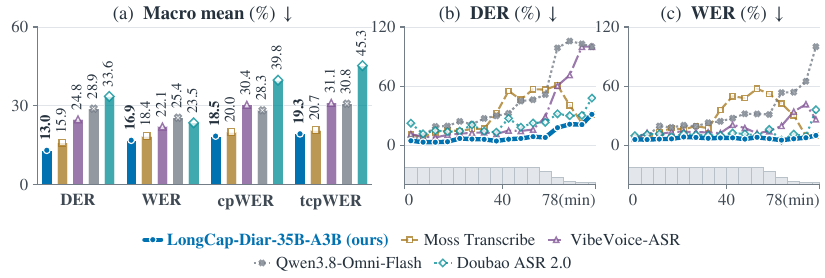}
\caption{Speech recognition and diarization:
(a) DER, WER, cpWER, and tcpWER, macro-averaged over five benchmark conditions;
(b--c) LongCapBench-Diar-Long errors in disjoint five-minute windows.
Gray bars count contributing recordings. Benchmark-specific
checkpoints and AMI overlap are described in App.~\ref{app:diar_results}.}
\label{fig:diar_long}
\vspace{-6pt}
\end{figure}

\subsection{Diarization Results}
\label{ssec:exp_diar}

\begin{figure}[!htbp]
\centering
\includegraphics[width=\linewidth]{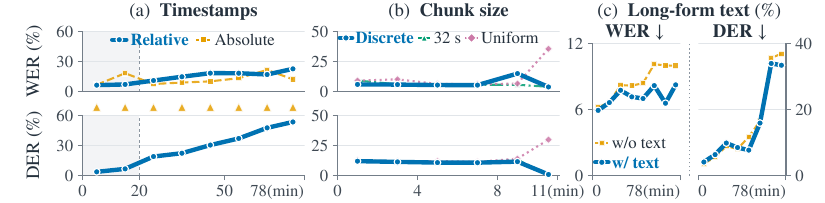}
\caption{Diarization design comparisons:
(a) Relative vs. absolute timestamps.
(b) Different chunking strategies.
(c) With and without long-form text in the training mixture.
Comparison protocols are detailed in App.~\ref{app:diar_results}.}
\label{fig:diar_ablation}
\vspace{-6pt}
\end{figure}

\textbf{Overall performance.}
LongCap-Diar-35B-A3B achieves the lowest macro-averaged DER, WER,
cpWER, and tcpWER in Fig.~\ref{fig:diar_long}a.
We further study temporal robustness on LongCapBench-Diar-Long, whose
recordings span 50--78 minutes (Fig.~\ref{fig:diar_long}b--c).
In five-minute windows, WER remains below 10\% throughout, while DER
stays below 9\% through 60 minutes before rising in the sparsely sampled
tail. These results show that our chunk-wise model can outperform the
evaluated holistic baselines, MOSS Transcribe Diarize and VibeVoice-ASR
\citep{moss2026transcribe,peng2026vibevoiceasr}, as well as the commercial
Qwen3.8-Omni-Flash \citep{qwen38omni2026} and Doubao ASR 2.0
\citep{doubaoasr2026} APIs.
Holistic versus chunk-wise processing is further ablated
in Table~\ref{tab:chunkwise_long}; the absolute/relative timestamp
comparison is shown separately in Fig.~\ref{fig:diar_ablation}a.
Our model is also better at dealing with very long and far-field spontaneous
speech (CHiME-6). With a 1M-token context window, LongCap-Diar can further
extrapolate to 16 hours of concatenated audio. Details are in
\S\ref{app:diar_results}.

\textbf{Design comparisons.}
The timestamp and chunk-policy comparisons use 2B or 35B-A3B models
trained on reduced data, so their scores are not directly comparable to
the main results. In Fig.~\ref{fig:diar_ablation}a, relative timestamps
slightly worsen WER but avoid absolute timestamps' complete DER failure
beyond the 20-minute training span; the absolute-timestamp checkpoint
was selected by test WER among 15 checkpoints.
Panel (b) shows little sensitivity to chunk size.
In (c), the long-text condition has lower WER, but also uses a different
training mixture and larger token budget.

\FloatBarrier
\subsection{Timestamped Audio Caption Results}

\begin{figure}[!htbp]
\centering
\includegraphics[width=\linewidth]{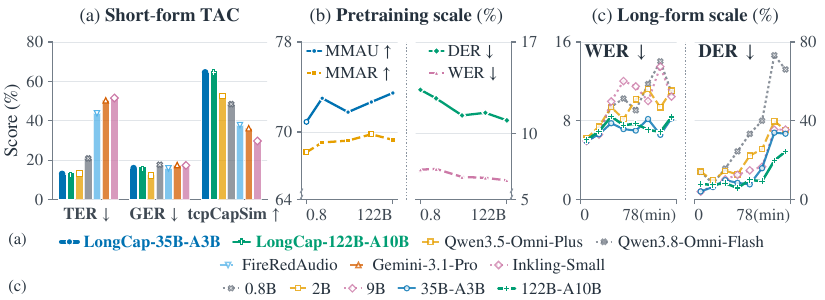}
\caption{(a) Short-form TAC results on LongCapBench-TAC-Short.
(b--c) Continued-pretraining scale comparisons: (b) MMAU/MMAR accuracy
and DER/WER averaged equally over English and Chinese evaluation sets;
(c) WER and DER over long-form recordings.}
\label{fig:caption_short_scaling}
\vspace{-6pt}
\end{figure}

\textbf{Overall Results}:
LongCap achieves strong TAC performance on both short clips and long
recordings. On LongCapBench-TAC-Short (Fig.~\ref{fig:caption_short_scaling}a),
both LongCap variants outperform the evaluated baselines in caption
similarity and achieve competitive timeline accuracy, while
Qwen3.5-Omni-Plus retains lower granularity error.
On LongCapBench-TAC-Long (Fig.~\ref{fig:teaser}), LongCap maintains higher
caption similarity and lower timeline error than the baselines in the
segment-level comparison. Full-recording measurements extend to 117.1
minutes, with higher timeline error in some later windows and substantial
music granularity errors in the unplotted final partial window. Coverage also decreases
from four recordings to one.\footnote{Missing events are already penalized
by TER and are therefore not penalized again by GER. On TAC-Long, most
semantic groups for some competitors contain no hypothesis events,
explaining their very low GER.}
A supplemental reference-grounded comparison with AF-Next and LAT using
native captioning prompts is reported in App.~\ref{app:native_caption_results}.

\begin{figure}[!htbp]
\centering
\includegraphics[width=\linewidth]{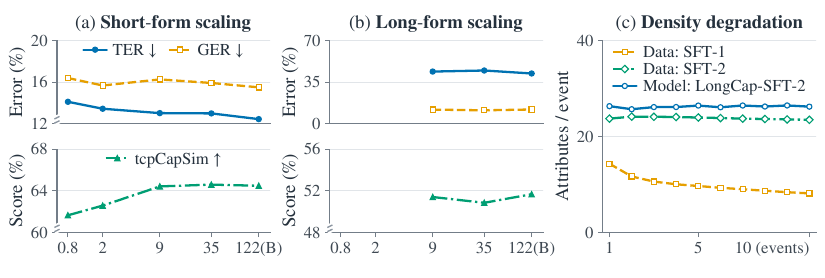}
\caption{TAC scaling and caption density.
(a--b) LongCap-SFT-2 on LongCapBench-TAC-Short and LongCapBench-TAC-Long;
0.8B and 2B lack reliable full-recording outputs.
(c) Filled attributes per event by position within the chunk.}
\label{fig:sft_analysis}
\vspace{-6pt}
\end{figure}

\textbf{Scaling Results}:
Continued pre-training yields modest gains on short-form audio tasks
(Fig.~\ref{fig:caption_short_scaling}b), but larger improvements in
long-form recognition and diarization (Fig.~\ref{fig:caption_short_scaling}c).
For post-training, Fig.~\ref{fig:sft_analysis}a--b compares
\texttt{LongCap-SFT-2} before data self-distillation.
Scaling from 0.8B/2B to 9B/35B-A3B lowers short-form TER and raises
tcpCapSim from 61.65/62.58\% to 64.42/64.59\%, while GER varies little
(Fig.~\ref{fig:sft_analysis}a).
The practical gain is greater on long-form audio: 0.8B and 2B produce
frequent invalid or repetitive captions far beyond the training range,
precluding reliable full-recording evaluation
(Fig.~\ref{fig:sft_analysis}b; App.~\ref{app:training_inference}).
With limited additional gains at 122B-A10B, 9B and 35B-A3B offer practical
model sizes for TAC.

\textbf{Density Degradation}:
Chunk-wise SFT-1 labels contain fewer filled attributes for later events
within each chunk (Fig.~\ref{fig:sft_analysis}c).
Event-level reannotation in SFT-2 addresses this degradation, producing
more consistent attribute density across event positions.
\texttt{LongCap-SFT-2} learns this improved profile, maintaining event-level
detail in its full-recording outputs.

\textbf{Long-form Expansion Results}:
The main benefit of long-form expansion is lower full-recording TER
(Table~\ref{tab:self_distillation}, App.~\ref{app:self_distillation_results}):
44.84\% to 41.09\% at 35B-A3B and 42.35\% to 39.64\% at 122B-A10B,
compared with smaller TER gains on short-form audio.
This pattern is consistent with extending supervision to longer recordings.
Long-form GER also decreases, while tcpCapSim changes are mixed.

\FloatBarrier
\begin{wrapfigure}{r}{0.40\linewidth}
\centering
\vspace{-20pt}
\includegraphics[width=\linewidth]{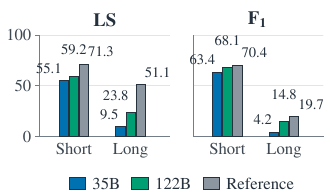}
\caption{LS and $F_1$ (\%) for 35B-A3B, 122B-A10B, and GLM-5.3-Flash on reference events.}
\label{fig:dependency}
\end{wrapfigure}
\textbf{Dependency Modeling}:
We evaluate predicted ancestor links against DeepSeek-V4-Pro and
Gemini-3.1-Pro silver annotations \citep{deepseek2026api,gemini31pro}.
Link support (LS) measures reference
support for individual links; connectivity $F_1$ measures recovery of
direct and transitive reference connections. Both models show substantial
gaps on long recordings (Fig.~\ref{fig:dependency}). Despite limited
captioning gains beyond 35B-A3B, 122B-A10B improves long-form connectivity
$F_1$ from 4.2\% to 14.8\%.
App.~\ref{app:dependency_results} details matching, checkpoints, and
GLM labeling controls.

\WFclear
\FloatBarrier
\section{Related Work}
\label{sec:related}

\textbf{Long-Form Speech Recognition and Diarization}: Earlier work studies
long-form generalization \citep{narayanan2019longform} and full-document
contexts \citep{chen2024trainlong}. Whisper decodes sequential 30-second
windows with window-relative timestamps \citep{radford2023whisper}.
CHiME benchmarks unsegmented recognition and diarization of distant speech
\citep{watanabe2020chime6,cornell2024chime8}. Recent speech LLMs jointly emit speaker
identities, timestamps, and transcripts
\citep{moss2026transcribe,peng2026vibevoiceasr,mu2026diascriber}; MOSS and
VibeVoice-ASR support single-pass inputs of up to 90 and 60 minutes,
respectively. Chunk-wise decoding extends from ASR
\citep{jia2024speechllmxl} to joint transcription and diarization with
cross-chunk speaker caches \citep{shi2025jedis,peng2026gstar}.
LongCap extends this approach to structured TAC and validates length
extrapolation in recognition and diarization (\S\ref{ssec:exp_diar}).

\textbf{Audio Captioning}: Audio captioning classically pairs a clip of
ten to thirty seconds with one free-text sentence
\citep{kim2019audiocaps,drossos2019clotho}.
Bagpiper uses rich captions for open-ended audio tasks \citep{tian2026bagpiper}.
Omni-Captioner studies the tradeoff between descriptive detail and
hallucination \citep{ma2026omnicaptioner}; the
Qwen3-Omni-Captioner model card advises inputs of at most 30 seconds
\citep{qwen3captioner2025}. Timestamped variants attach start and end
times to each described event, echoing dense video captioning
\citep{krishna2017densecap}. The TAC model \citep{kumar2026tac}, distinct
from our task acronym, emits typed, timestamped descriptions; its TAC-V
pipeline processes independent 20-second audio chunks.
AudioMap names the task time-aware dense audio captioning
\citep{rong2026audiomap}, AudioTime supplies temporally aligned training
data \citep{xie2025audiotime}, LAT reaches 30-minute inputs and
contributes the human-verified LAT-Bench \citep{shao2026lataudio}, and
audio diarization localizes open-set events without describing them
\citep{werning2026audiodiarization}. Hour-scale inputs have been reached
for question answering and temporal grounding \citep{kutsakov2026gigachat}.
LongCap combines retained context across chunks (\S\ref{ssec:chunkwise}),
62-key structured captions (\S\ref{ssec:formulation}), and long-form
training (\S\ref{ssec:sft}) with human-labeled LongCapBench evaluation
approaching two hours (\S\ref{ssec:benchmark}).

\section{Conclusion}
\label{sec:conclusion}

LongCap enables hour-scale TAC through chunk-wise generation, relative
timestamps, and detailed event supervision. On LongCapBench-TAC, 9B and
35B-A3B offer favorable size--performance tradeoffs, while 122B-A10B
better recovers event dependencies. This distinction highlights different
scaling needs for describing events and connecting them over time,
providing a foundation for long-horizon audio understanding and
always-on voice agents.

\subsection*{AI Use Statement}
We used generative AI extensively for data construction, annotation,
refinement, and quality assessment. We also used LLMs for event grouping
and semantic-similarity evaluation, as described in
\S\ref{ssec:metrics}. AI assistants supported portions of code development
and experiment management, and helped polish the manuscript and
proofread its text. The core research ideas were conceived entirely
by the human researchers.

\subsection*{Ethics Statement}
Our data preparation uses public and private audio resources and
model-assisted annotation; human annotators finalize the benchmark labels.
Our collaboration with audio professionals and human annotators followed
standard ethical practices and respected their rights, welfare, and privacy.
Speech and speaker attributes can expose sensitive information, and
generated descriptions may contain errors or reflect biases in the source
data and annotation models. These limitations matter for applications
involving speaker profiling or consequential decisions.
The four LongCapBench-TAC-Long recordings have research-use permission;
this does not by itself establish permission to redistribute their audio.

\subsection*{Reproducibility Statement}
Appendix~\ref{app:scoring_benchmark} specifies the attribute schema,
scoring rules, and benchmark composition. Appendix~\ref{app:setup}
describes data curation, training configurations, and inference safeguards,
and Appendix~\ref{app:prompts} provides the seed and event-wise annotation
prompts. Appendix~\ref{app:additional_results} documents additional
evaluation protocols, checkpoint comparisons, and their limitations.

\subsection*{Acknowledgments}
We thank Xinying Hu, Shuangshuang He, and Zhuoqun Liu for their expert
annotation work.
We thank Anuttacon for its support in this work.

\bibliography{iclr2027_conference}
\bibliographystyle{iclr2027_conference}

\clearpage
\appendix
\raggedbottom
\section{Scoring and Benchmark}
\label{app:scoring_benchmark}

\subsection{Attribute Key Set}
\label{app:taxonomy}

The structured caption schema contains 62 unique keys organized into six
top-level categories. The tree below presents the complete key set in
English. Each leaf is marked by the event types to which it applies:
\texttt{[S]} for speech, \texttt{[M]} for music, and \texttt{[E]} for sound
events. Speech, music, and sound events draw on 33, 25, and 9 keys,
respectively; shared keys appear once with multiple marks. A key is
included in an event's caption only when applicable.
The numeric prefixes retain identifiers from the broader annotation
taxonomy rather than defining a consecutive list. Subcategories 1.3,
3.2, 4.2, 6.3, and 6.4 have no retained keys in this event-level schema
and are intentionally absent; the displayed 62-key set is complete.

\begingroup
\let\footnotesize\scriptsize
\begin{framed}
\footnotesize
\begin{verbatim}
1. Source Identity & Inventory
|-- 1.1 Human Voice Sources
|   |-- speaker identity                                 [S]
|   |-- speaker age                                      [S]
|   |-- speaker gender                                   [S]
|   `-- vocal role type                                  [S]
|-- 1.2 Musical Sources
|   |-- instrument list                                  [M]
|   |-- vocal type                                       [M]
|   `-- instrument role                                  [M]
`-- 1.4 Source Provenance & Authenticity
    |-- source authenticity                              [E]
    `-- edit / splice detection                          [E]

2. Linguistic & Verbal Content
|-- 2.1 Transcription
|   |-- orthographic transcription                       [S]
|   `-- lyrics transcription                             [M]
|-- 2.2 Language & Dialect
|   |-- language identification                          [S]
|   |-- dialect                                          [S]
|   |-- accent                                           [S]
|   `-- code-switching                                   [S]
|-- 2.3 Lexical & Grammatical Properties
|   |-- vocabulary level                                 [S]
|   |-- syntactic complexity                             [S]
|   `-- named entities                                   [S]
`-- 2.4 Disfluency & Non-lexical Vocalizations
    |-- disfluency types                                 [S]
    |-- non-lexical vocalizations                        [S]
    `-- backchannel tokens                               [S]

3. Acoustic & Signal Properties
|-- 3.1 Loudness & Dynamics
|   `-- loudness contour                                 [S,E]
|-- 3.3 Timbre & Texture
|   |-- voice quality                                    [S]
|   |-- vocal acoustic detail                            [S]
|   |-- instrument timbre descriptors                    [M]
|   `-- timbre & texture (subcategory asked as one key)  [E]
|-- 3.4 Speech Prosody
|   |-- speech rate                                      [S]
|   |-- stress & prominence                              [S]
|   |-- pause distribution                               [S]
|   `-- pitch and intonation                             [S]
|-- 3.5 Musical Acoustic Properties
|   |-- tempo / bpm                                      [M]
|   |-- beat strength                                    [M]
|   |-- key / tonality                                   [M]
|   |-- chord progression                                [M]
|   |-- main melody                                      [M]
|   |-- melody contour                                   [M]
|   |-- key modulation / tonal changes                   [M]
|   `-- musical texture type                             [M]
|-- 3.6 Singing Voice Attributes
|   |-- vocal technique                                  [M]
|   |-- singing timbre                                   [M]
|   `-- dynamics in singing                              [M]
|-- 3.7 Recording & Signal Quality
|   `-- perceived audio quality                          [S]
|-- 3.8 Production & Sound Design
|   `-- intentional effects                              [M]
`-- 3.9 Instrumental Performance Attributes
    |-- instrumental articulation                        [M]
    `-- playing dynamics                                 [M]
\end{verbatim}

\begin{verbatim}
4. Temporal Structure
|-- 4.1 Segmentation & Boundaries
|   `-- music structure sections                         [M]
`-- 4.3 Temporal Relationships & Dynamics
    |-- tempo changes / rhythmic evolution               [M]
    `-- turn-taking patterns                             [S]

5. Spatial & Environmental Properties
|-- 5.1 Room & Environmental Acoustics
|   |-- reverberation character                          [M,E]
|   `-- distance estimation                              [E]
|-- 5.2 Noise Environment
|   `-- foreground-background separation                 [E]
`-- 5.3 Environmental & Event Sound Sources
    `-- sound event identification                       [E]

6. Semantic & Pragmatic & Cultural Interpretation
|-- 6.1 Emotion & Affect
|   |-- speaker emotion                                  [S]
|   |-- emotional intensity / arousal                    [S,M]
|   |-- valence trajectory                               [S]
|   `-- music mood                                       [M]
|-- 6.2 Intent & Pragmatics
|   |-- speaker intent                                   [S]
|   |-- speech act type                                  [S]
|   |-- subtext                                          [S]
|   `-- rhetorical strategy                              [S]
|-- 6.5 Cultural, Domain & Contextual Knowledge
|   `-- cultural context                                 [S]
`-- 6.6 Cross-modal Interaction
    `-- sound function                                   [S,M,E]
\end{verbatim}
\normalsize

\end{framed}
\endgroup

\pagebreak
\subsection{Scoring Rules}
\label{app:metric_formula}

\textbf{Diarization Error Rate (DER)}:
We use \texttt{pyannote.metrics} \citep{bredin2017pyannotemetrics}
with optimal one-to-one speaker assignment. DER \citep{fiscus2007rt07} divides missed-speech,
false-alarm, and speaker-confusion duration by reference speaker duration.
Overlapping speech is included, counting simultaneous speakers separately.
The 0.25\,s boundary collar excludes 0.125\,s on either side of each
reference boundary.

\textbf{cpWER and tcpWER}:
MeetEval \citep{vonneumann2023meeteval} concatenates each speaker's
utterances chronologically and finds the speaker permutation minimizing
substitutions, deletions, and insertions. cpWER divides this count by
reference transcript length; tcpWER restricts word alignment using
estimated word timings, extending hypothesis intervals by 5\,s on each
side. This allowance permits temporal matching without excluding boundary
regions. Both transcripts are normalized, scoring English
by word and Chinese by character. Speaker-agnostic English WER uses Whisper's
\texttt{EnglishTextNormalizer} \citep{radford2023whisper}.\footnote{LongCapBench cpWER/tcpWER use
\href{https://github.com/wenet-e2e/WeTextProcessing}{WeTextProcessing}'s English/Chinese normalizers; CHiME-6 uses CHiME8
normalization.}

\textbf{TAC event grouping}:
DER and tcpWER optimize over speaker permutations, using speaker labels
to define utterance groups. General audio events lack equivalent identity
labels, so we approximate enumeration of possible groupings with three
proposals from DeepSeek-V4-Flash \citep{deepseek2026api} with thinking enabled.
Using the event intervals $[t_i^s,t_i^e)$ defined in
\S\ref{sec:task}, each proposal groups reference and predicted
events describing the same occurrence. Each event belongs to one group,
with multiple events or an empty side allowed. Repeated occurrences from
the same speaker or sound source remain distinct. For each proposal,
grouping advances in 60\,s steps with a 15\,s look-back, retaining full
intervals and merging groups that share events across windows into a
recording-level grouping. Each metric independently retains its best
result across the three candidates: the lowest TER, the lowest GER, or
the highest tcpCapSim (\S\ref{ssec:metrics}).

\textbf{TER}:
TER applies DER accounting to semantic groups. Merge overlapping intervals
within each group separately on each side. Too few active predicted
groups count as misses, too many as false alarms, and remaining activity
in different groups as confusion. Divide these error durations by the
merged reference duration $T_\mathrm{ref}$ in Eq.~(\ref{eq:ter}). Overlap
between different groups counts separately. TER and GER use full intervals
without a boundary collar.

\textbf{GER}:
GER compares how shared activity is divided into individual events.
One-to-one matching permits each event only one counterpart, so splitting
or merging can reduce matched overlap even when time coverage is unchanged.
In Eq.~(\ref{eq:ter}), $T_\mathrm{overlap}$ sums the overlap of merged
reference and predicted activity within groups; $T_\mathrm{match}$ is the
maximum total overlap under one-to-one event matching within groups;
and $T_\mathrm{dup}$ is summed prediction duration minus merged prediction
duration within groups. Divide
$T_\mathrm{overlap}-T_\mathrm{match}+T_\mathrm{dup}$ by the total reference
event duration $\sum_i(t_i^e-t_i^s)$, counting overlapping events separately.
For example, splitting a reference event $[0,10)$ into predictions
$[0,5)$ and $[5,10)$ in one group preserves all 10 seconds of coverage,
but only one 5-second piece can match the reference event. Thus TER $=0$
while GER $=(10-5)/10=0.5$. Merging these two reference events into one
prediction gives the same penalty. Predicting $[0,10)$ twice for a single
reference event instead adds 10 seconds through $T_\mathrm{dup}$, giving
GER $=1$ despite unchanged coverage.

\textbf{tcpCapSim}:
Pool nonempty caption values by key within each group, counting each key
once per side. Scoring includes \texttt{type} and excludes
free-text \texttt{caption} and structural \texttt{speaker\_id}. A group
receives credit only if some reference and prediction intervals overlap
or are at most 5\,s apart. This matching allowance does not exclude
boundary regions. DeepSeek-V4-Flash scores each shared key's semantic
agreement in $\{0,0.25,0.5,0.75,1\}$; missing counterparts and groups
failing the time constraint receive zero. Summed credit divided by
predicted and reference key counts gives precision and recall.
tcpCapSim is their arithmetic mean, as described in \S\ref{ssec:metrics}.

\textbf{Aggregation}:
All metrics use micro aggregation within each benchmark: sum error counts
or durations, similarity credit, and their denominators across recordings
before computing rates. Reported cross-benchmark macro averages weight
benchmark scores equally.

\subsection{LongCapBench Details}
\label{app:benchmark}

LongCapBench intentionally covers both English and Chinese in its
short- and long-form subsets for diarization and TAC. We exclude every
LongCapBench source recording in its entirety from our training corpus.
Table~\ref{tab:longcapbench} summarizes the subset sizes and durations.

\textbf{TAC annotation workflow}:
Starting from Qwen3.5-Omni-Plus \citep{qwen35omni2026} predictions, annotators first correct event
timestamps, then verify each event's attributes individually. Model
proposals guide event discovery. We assign related attributes to specialized
annotator groups, and data batches pass through these groups, each checking
a small set of related attributes. Specialist review, particularly for
music, is a throughput bottleneck. Maintaining consistency across events
and resolving ambiguous annotation rules introduce further difficulties.

\textbf{LongCapBench-Diar-Short}:
The short-form diarization subset covers movie excerpts, meetings,
podcasts, and casual conversations involving multiple speakers. These
sources provide both scripted and spontaneous speech with varied
speaker-turn patterns.

\textbf{LongCapBench-Diar-Long}:
The long-form diarization subset contains 15 recordings, primarily
meetings lasting approximately 50--78 minutes, with 7--19 speakers per
recording and a mean of 9.7 speakers. It evaluates sustained transcription
and speaker tracking over long conversations.

\textbf{LongCapBench-TAC-Short}:
The short-form TAC subset covers speech, music, and sound effects. Music
includes both instrumental and vocal content. Movie clips provide
representative mixtures of these sound types within the same scene.

\textbf{LongCapBench-TAC-Long}:
The long-form TAC subset contains two movies and two shows, with one
movie and one show in each of English and Chinese. The movies
provide extended mixtures of speech, music, and sound effects, while the
shows provide long stretches of conversational and performed
speech. All four recordings have licenses permitting research use.
Table~\ref{tab:tac_long_sources} lists the recordings and their source
documentation.

\begin{table}[!htbp]
\caption{Sources and licensing documentation for LongCapBench-TAC-Long}
\label{tab:tac_long_sources}
\centering
\small
\setlength{\tabcolsep}{4pt}
\renewcommand{\arraystretch}{1.15}
\scalebox{0.94}{%
\begin{tabular}{@{}p{0.30\linewidth}lrl@{}}
\toprule
Recording & Language & Min. & Source \\
\midrule
\emph{A Star Is Born} (1937)
  & English & 110.67
  & \href{https://s3.amazonaws.com/ava-dataset/trainval/HKjR70GCRPE.mp4}{link} \\
\addlinespace[3pt]
\emph{Personal Tailor} (2013)
  & Chinese & 117.05
  & \href{https://huggingface.co/datasets/sy1998/MLVU/resolve/main/MLVU/video/sub_scene/subPlot_new_all_166.mp4}{link} \\
\addlinespace[3pt]
92nd Oscars (2020)
  & English & 93.60
  & \href{https://www.youtube.com/watch?v=-WnyRMZqV1U}{link} \\
\addlinespace[3pt]
\emph{Roast!} (S3E05)
  & Chinese & 101.49
  & \href{https://www.youtube.com/watch?v=5zcD0IpLc-Q}{link} \\
\bottomrule
\end{tabular}%
}
\end{table}

\clearpage
\section{Implementation}
\label{app:setup}

\subsection{Data Curation}
\label{app:data}

The initial pre-training and SFT recipe uses audio examples of at most
20 minutes, including examples assembled by splicing. Optional length
expansion extends the maximum duration to 40 minutes.

\subsubsection{Pre-training Data}

\textbf{Diarization data}:
We curate speech-only recordings from web resources and use a cascaded
pipeline for automatic annotation and quality filtering. To reduce
speaker-clustering degradation on long recordings, we first split the
audio into recordings of at most 20 minutes and identify their languages
with Whisper \citep{radford2023whisper}. Paraformer \citep{gao2022paraformer} and Whisper
large-v3-turbo perform the initial transcription of Chinese and English
audio, respectively, producing text and timestamps. \mbox{WeSpeaker} \citep{wang2023wespeaker} extracts
speaker embeddings, which are clustered to assign utterance-level speaker
labels. We then refine utterance boundaries using temporal alignment and
local energy information to reduce the risk of clipping speech. Finally,
Qwen3-ASR \citep{qwen3asr2026} performs a second transcription pass, whose output provides the
final transcript. We filter low-quality recordings based on agreement
between the two transcription passes, the proportion of speech, and the
token rate.

\textbf{Clip-level natural-language captions}:
We curate audio from private resources and publicly available datasets,
covering speech, non-speech sounds, and their mixtures. The audio is
semantically segmented into chunks of up to 30 seconds, and
Qwen3-Omni-Captioner \citep{qwen3captioner2025} generates a natural-language
caption for each chunk. We apply lightweight filtering to remove obvious
labeling failures, such as repetitive text loops. After chunk-level
annotation, we splice the chunks and their captions into training
recordings with an overall duration of up to 10 minutes. Speech
constitutes the majority of this corpus; non-speech content is harder to
obtain and accounts for a smaller share.

\textbf{Long-form text}:
We use the Dolma3 Longmino Mix corpus for OLMo-3 \citep{olmo2025olmo3}.\footnote{Corpus identifier:
\href{https://huggingface.co/datasets/allenai/dolma3_longmino_mix-100B-1125}{\texttt{allenai/dolma3\_longmino\_mix-100B-1125}}.}
Its original examples contain at most 64k tokens. We concatenate examples
from the same domain to construct training contexts of up to 256k tokens.
We sample across short and long sequences to keep the sequence-length
distribution approximately uniform.

\subsubsection{Supervised Fine-Tuning Data}

The SFT audio follows the same broad source and content distribution as
the captioning pre-training data, with additional attention to balancing
coverage across speech, music, sound effects, and their mixtures. The two
initial stages in \S\ref{ssec:sft} use 15k and 10k hours of audio retained
after filtering for seed supervision and event-wise reannotation,
respectively. These recordings average 49 seconds and last at most
10 minutes. Optional length expansion adds 20k hours of audio with broad
acoustic coverage, including approximately 10k hours from recordings
longer than 20 minutes, with a maximum duration of 40 minutes.
All of this SFT audio is held out from the pre-training corpus.
The seed and event-wise annotation prompts are provided in
App.~\ref{app:prompts}.

\subsection{Training and Inference Details}
\label{app:training_inference}

\textbf{Hardware and data mixture}:
Our diarization models are pre-trained on 32 H100 GPUs. TAC models use 128
H100 GPUs.
Each GPU processes 32k tokens per step, corresponding to 1M tokens per
step for diarization and 4M for TAC. Continued pre-training lasts 100k steps.
The captioning:diarization:text mixture is $1:1:1$ during pre-training
and $3:1:1$ during SFT, measured by token count.

\textbf{Training setup}:
Table~\ref{tab:training_setup} summarizes the pre-training configuration.
Context parallelism of degree 8 distributes each
sequence across eight GPUs, supporting contexts of up to 256k tokens.
SFT uses the same training configuration with the $3:1:1$ data mixture
above and a constant learning rate of $1\times10^{-5}$ for two epochs
over the retained SFT data.

\begin{table}[!htbp]
\caption{Pre-training configuration. The batch size shown is for TAC training
on 128 GPUs; diarization uses 32 GPUs.}
\label{tab:training_setup}
\centering
\small
\setlength{\tabcolsep}{6pt}
\renewcommand{\arraystretch}{1.12}
\scalebox{0.94}{%
\begin{tabular}{@{}ll@{}}
\toprule
Setting & Value \\
\midrule
Optimizer & AdamW \\
Pre-training steps & 100k \\
Pre-training warmup & 2k steps \\
Pre-training peak learning rate & $2\times10^{-5}$ \\
Pre-training learning-rate decay & Cosine decay to $1\times10^{-5}$ \\
Tokens per GPU & 32k \\
TAC batch size & $32\mathrm{k}\times128$ tokens ($\approx$4M tokens/step) \\
Context parallelism & 8 \\
Maximum training context & 256k tokens \\
Maximum pre-training audio duration & 20 minutes \\
\bottomrule
\end{tabular}%
}
\end{table}

\textbf{Inference failures}:
LongCap uses greedy decoding with a maximum of 8,192 output tokens per
chunk. If a chunk reaches this limit without completing its caption,
subsequent chunks are still decoded normally. For the 35B-A3B model,
approximately 0.3--0.5\% of chunks trigger a retry because of malformed
captions, such as unparseable JSON or loops that repeatedly emit events.
These failures become more frequent with smaller models and longer
contexts, making the 0.8B and 2B variants impractical for reliable
long-form use.

\textbf{Retries and postprocessing}:
We use two measures to reduce these failures. First, when a malformed
caption is detected, we regenerate the same chunk up to three times,
using repetition penalties of 1.05, 1.10, and 1.15 for successive retries.
Every retry still uses greedy decoding. Second, after decoding, we remove
obviously repetitive duplicate events, such as a run of more than 100
events repeating the same transcript.

\textbf{Caption-based audio understanding}:
On MMAU and MMAR, we evaluate how well LongCap's captions support
question answering with a text LLM. LongCap first captions each recording;
\texttt{gemini-3-flash-preview} \citep{gemini3flash} then answers the question using the caption
and the supplied answer choices, without access to the audio.
The reported accuracy measures this two-stage pipeline, rather than
LongCap answering the questions directly. This follows the evaluation
practice described in \S\ref{ssec:setup}.

\clearpage
\section{Prompts}
\label{app:prompts}
\label{app:rubrics}

\subsection{Prompts}
\label{app:prompt_inventory}

\textbf{Prompt 1: Seed structured annotation}:
\label{prompt:seed}
\begin{promptbox}
\fontsize{7}{8.5}\selectfont
\setlength{\topsep}{0pt}\setlength{\partopsep}{0pt}
\begin{alltt}
Listen to the audio. Output one JSON object (no markdown). Schema:

\{
"timeline":[\{"time\_stamps":["<Xs>","<Ys>"],"type":"speech|music|sound\_event",
    "speaker\_id":"S1|null","transcript":"<text>|null",
    "attributes":\{"<id> <attr>":"<val>",...\},"caption":"<Chinese>"\}]\}

TIMELINE (select relevant per segment):
<attribute keys from Appendix \ref{app:taxonomy}, selected by event type>
speech: [S]; music: [M]; sound\_event: [E].

RULES:
1. Concurrent layers (music+speech) -> separate overlapping timeline
entries.
2. Speaker IDs S1,S2,... assigned consistently; null for non-speech.
3. transcript = exact spoken words for speech; null otherwise.
4. All values and captions in Chinese. Arrays for list values.
5. time\_stamps as float seconds: "<X.Xs>","<Y.Ys>".
\end{alltt}
\end{promptbox}

\textbf{Prompt 2: Event-wise reannotation}:
\label{prompt:local}
\begin{promptbox}
\fontsize{7}{8.5}\selectfont
\setlength{\topsep}{0pt}\setlength{\partopsep}{0pt}
\begin{alltt}
Listen to this short audio segment only. The segment type has already been
identified as "<event\_type>". Produce LOCAL structured annotation for this
single segment only.

Output one JSON object (no markdown) with exactly this schema:
\{
  "type": "<event\_type>",
  "time\_stamps": ["0.0s", "<duration to two decimals>s"],
  "speaker\_id": "S1|null",
  "transcript": "<exact spoken words or lyrics>|null",
  "attributes": \{"<id> <attr>": "<value>|null", ...\},
  "caption": "<Chinese>"
\}

Annotate only these attributes for this segment type. Include every key; use
null when the attribute is not present or cannot be inferred from the audio:
<all keys for event\_type from Appendix \ref{app:taxonomy}>

Existing segment text for context only. Re-listen to the audio and correct
it when the audio disagrees:
<existing event text, or "(empty)">

Rules:
1. Do not output timeline.
2. Keep the output type as "<event\_type>".
3. speaker\_id is only for speech; use null for music and sound\_event.
4. transcript is exact spoken words for speech, exact lyrics for music if
any, and null for non-vocal music or sound\_event.
5. All attribute values and captions must be in Chinese.
6. time\_stamps are relative to this sliced segment, from 0.0s to its
duration.
\end{alltt}
\end{promptbox}

\clearpage
\section{Additional Results}
\label{app:additional_results}

\subsection{Additional Diarization Results}
\label{app:diar_results}

\begin{table}[!htbp]
\caption{Speech recognition and diarization results on each benchmark.
Each cell reports DER~/~WER on the first line and cpWER~/~tcpWER on the
second, in \%; lower is better.}
\label{tab:diar_public}
\centering

\normalsize
\setlength{\tabcolsep}{4pt}
\newcommand{\diarcell}[2]{\begin{tabular}[c]{@{}c@{}}#1\\#2\end{tabular}}
\scalebox{0.94}{%
\begin{tabular}{@{}lccccc@{}}
\toprule
Test set & MOSS & VibeVoice-ASR & \shortstack{Doubao ASR\\2.0}
         & \shortstack{Qwen3.8-\\Omni-Flash} & \shortstack{LongCap-Diar\\(ours)} \\
\midrule
AliMeeting & \diarcell{\textbf{9.27}/\textbf{18.30}}{\textbf{15.49}/\textbf{15.68}} & \diarcell{34.73/35.99}{38.28/38.62} & \diarcell{28.63/24.65}{32.13/34.57} & \diarcell{18.60/24.20}{21.92/22.26} & \diarcell{9.49/20.46}{16.30/16.65} \\
\addlinespace[2pt]
AISHELL-4 & \diarcell{\textbf{5.80}/13.44}{\textbf{13.15}/\textbf{13.43}} & \diarcell{13.57/18.19}{22.29/22.70} & \diarcell{24.66/\textbf{12.59}}{34.46/41.48} & \diarcell{10.19/14.58}{15.40/15.73} & \diarcell{8.37/14.26}{16.94/17.44} \\
\addlinespace[2pt]
AMI-IHM & \diarcell{25.17/24.61}{\textbf{20.92}/\textbf{21.58}} & \diarcell{27.18/22.50}{24.35/25.13} & \diarcell{47.15/31.81}{49.29/56.40} & \diarcell{47.84/40.57}{37.01/39.32} & \diarcell{\textbf{21.98}/\textbf{21.03}}{21.70/22.46} \\
\addlinespace[2pt]
AMI-SDM & \diarcell{27.97/29.87}{32.95/34.09} & \diarcell{31.60/27.58}{38.75/39.73} & \diarcell{50.37/42.91}{57.41/66.56} & \diarcell{44.69/35.56}{35.83/41.14} & \diarcell{\textbf{15.46}/\textbf{22.73}}{\textbf{21.86}/\textbf{23.48}} \\
\midrule
LongCapBench-Diar-Short & \diarcell{11.41/5.98}{17.34/18.84} & \diarcell{16.96/6.28}{28.24/29.48} & \diarcell{17.00/\textbf{5.52}}{25.49/27.51} & \diarcell{23.18/12.31}{31.10/35.48} & \diarcell{\textbf{9.94}/5.89}{\textbf{15.47}/\textbf{16.56}} \\
\addlinespace[2pt]
LongCapBench-Diar-Long & \diarcell{27.11/27.00}{31.77/32.63} & \diarcell{15.45/14.23}{20.35/25.69} & \diarcell{18.05/10.74}{18.54/20.73} & \diarcell{29.65/23.52}{28.38/43.72} & \diarcell{\textbf{6.02}/\textbf{6.85}}{\textbf{13.13}/\textbf{14.06}} \\
\midrule
\textbf{Macro mean} & \diarcell{17.79/19.87}{21.94/22.71} & \diarcell{23.25/20.80}{28.71/30.23} & \diarcell{30.98/21.37}{36.22/41.21} & \diarcell{29.03/25.12}{28.27/32.94} & \diarcell{\textbf{11.88}/\textbf{15.20}}{\textbf{17.57}/\textbf{18.44}} \\
\bottomrule
\end{tabular}%
}
\end{table}

\begin{table}[!htbp]
\caption{CHiME-6 results on sessions S01, S02, S09, and S21, micro-averaged
across sessions. The test set has multiple near-field recordings for each example. We infer them all and keep the best results of each example. The maximum audio length is 2 hours 40 minutes. Most competitors are not practically usable on this application.}
\label{tab:chime6}
\centering
\normalsize
\setlength{\tabcolsep}{8pt}
\scalebox{0.94}{%
\begin{tabular}{@{}lrrrr@{}}
\toprule
Model & DER & WER & cpWER & tcpWER \\
\midrule
MOSS & 99.81 & 99.81 & 99.81 & 99.97 \\
VibeVoice-ASR & 100.00 & 99.99 & 99.99 & 100.00 \\
Doubao ASR 2.0 & 69.90 & 53.77 & 69.48 & 77.04 \\
Qwen3.8-Omni-Flash & 79.07 & 69.25 & 71.35 & 75.59 \\
LongCap-Diar (ours) & \textbf{29.04} & \textbf{37.08} & \textbf{32.29} & \textbf{34.14} \\
\bottomrule
\end{tabular}%
}
\end{table}

\begin{table}[!htbp]
\caption{
Comparison of 2B models with holistic processing and chunk-wise processing
on LongCapBench-Diar-Long.
All metrics are reported in \%; lower is better.
tcpWER uses a 5-second collar.
Best values are bold.
}
\label{tab:chunkwise_long}
\centering
\normalsize
\setlength{\tabcolsep}{6pt}
\renewcommand{\arraystretch}{1.1}

\begin{tabular}{@{}lrrrr@{}}
\toprule
Model & DER & WER & cpWER & tcpWER \\
\midrule
holistic processing
& 109.13 & 80.59 & 115.87 & 147.16 \\

chunk-wise processing
& \textbf{20.30}
& \textbf{12.74}
& \textbf{29.01}
& \textbf{34.80} \\
\bottomrule
\end{tabular}
\end{table}

\begin{figure}[!htbp]
\centering
\includegraphics[width=0.70\linewidth]{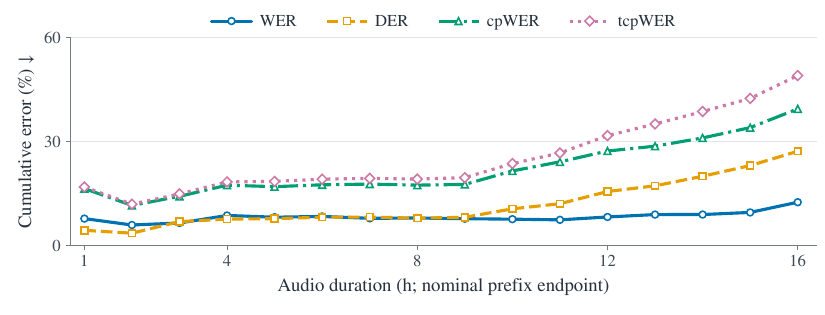}
\caption{Cumulative error rates up to 16 hours, averaged over three recording orders.}
\label{fig:diar_16h}
\vspace{-6pt}
\end{figure}

\textbf{Extrapolation to 16 hours:}
With a 1M-token context window, LongCap-Diar processes 16 hours of
concatenated audio. We average results over three recording-order
permutations (Fig.~\ref{fig:diar_16h}). WER remains stable throughout,
whereas DER, cpWER, and tcpWER increase because the speaker count in the
concatenated audio far exceeds the range seen during training.
This experiment is intended only to demonstrate long-duration processing.

\FloatBarrier

\subsection{Caption Content Comparison with AF-Next and LAT}
\label{app:native_caption_results}

We compare caption content on all 1,500 LongCapBench-TAC-Short clips
(569 source recordings). Following the segment-level inference setup in Fig.~\ref{fig:teaser},
we also evaluate these models on LongCapBench-TAC-Long using non-overlapping
10-minute chunks (45 chunks from four recordings, including shorter tails).
We use the main LongCap
checkpoints, AF-Next Captioner~\citep{ghosh2026audioflamingonext}, and
LAT-Audio-Base~\citep{shao2026lataudio}. AF-Next receives the
\href{https://huggingface.co/nvidia/audio-flamingo-next-captioner-hf/blob/62aa63a66927f04f8308620f8e4b65696d3ea0a6/README.md}{official captioning prompt}:
\emph{``Generate a detailed caption for the input audio. In the caption,
transcribe all spoken content by all speakers precisely.''}
LAT uses its \href{https://huggingface.co/datasets/mcshao/LAT-Bench/blob/30df66a85826c23c108f9fda03409e4b7aa18a01/task/bench-EN-DAC.jsonl}{released dense-captioning prompt}:
\emph{``Provide a dense audio description for this audio segment.''}
We decode greedily and set a 16,384-token output limit for both baselines,
using BF16 and repetition penalty 1.2 for AF-Next, and dynamic FP8 for LAT.

DeepSeek-V4-Pro jointly receives the reference and four anonymous candidates,
returning preferences or ties for all six pairs. Structured references and
LongCap outputs are deterministically rendered as field--value lists;
native baseline text is unchanged. No LLM paraphrases inputs. The rubric
balances salient content coverage against explicit reference contradictions,
ignoring schema, numerical timestamps, exact transcript matching, and style.
Unannotated details remain unknown. Two separate requests use a deterministic shuffled order and its reversal,
without sharing judgments. Inputs are untruncated; the output/reasoning
budget is 65,536 tokens.

For each LongCap--baseline pair, a win, tie, or loss receives 1, 0.5, or 0.
We average both orders, then inputs; Short's five equally sized subsets and
Long's chunks receive equal weight. Table~\ref{tab:native_caption_preference}
reports preference percentages; 50\% denotes equal preference credit.

Short's 95\% percentile intervals use 2,000 global source-cluster bootstrap
draws, sharing each source's multiplicity across subsets. These intervals describe
sensitivity to source composition. Long has only four source recordings.
Reference incompleteness, recognizable styles, and single-judge bias remain;
preferences do not measure timing or audio-verified accuracy.

\begin{table}[!htbp]
\caption{Tie-adjusted LongCap preference (\%), with 95\% source-cluster
bootstrap intervals for Short.}
\label{tab:native_caption_preference}
\centering
\small
\setlength{\tabcolsep}{7pt}
\begin{tabular}{@{}llcc@{}}
\toprule
LongCap model & Baseline & Short (95\% interval) & \shortstack{Long\\(10-min chunks)} \\
\midrule
LongCap-35B-A3B & AF-Next & 88.30 (86.80--89.63) & 97.22 \\
                 & LAT & 98.97 (98.48--99.39) & 97.78 \\
\addlinespace[2pt]
LongCap-122B-A10B & AF-Next & 87.75 (86.21--89.19) & 97.22 \\
                  & LAT & 99.20 (98.78--99.54) & 98.89 \\
\bottomrule
\end{tabular}
\end{table}

\Needspace{18\baselineskip}
\subsection{Length Expansion via Self-Distillation}
\label{app:self_distillation_results}

\begin{table}[!htbp]
\caption{Before and after self-distillation on all 1,500 Short clips and
all four full Long recordings. All values are percentages; arrows
indicate preferred directions. Bold marks the better value within each model-size pair.}
\label{tab:self_distillation}
\centering
\small
\setlength{\tabcolsep}{4pt}
\renewcommand{\arraystretch}{1.12}
\sisetup{detect-all, table-format=2.2, table-number-alignment=center}
\begin{tabular*}{\linewidth}{@{\extracolsep{\fill}}ll *{6}{S}@{}}
\toprule
\multirow{2}{*}{Model} & \multirow{2}{*}{Checkpoint}
& \multicolumn{3}{c}{Short form} & \multicolumn{3}{c}{Full long form} \\
\cmidrule(lr){3-5}\cmidrule(l){6-8}
& & {TER $\downarrow$} & {GER $\downarrow$} & {tcpCapSim $\uparrow$}
& {TER $\downarrow$} & {GER $\downarrow$} & {tcpCapSim $\uparrow$} \\
\midrule
\multirow{2}{*}{35B} & Before  & 13.01 & \bfseries 15.93 & \bfseries 64.59 & 44.84 & 11.46 & 50.85 \\
 & After  & \bfseries 12.42 & 16.13 & 63.75 & \bfseries 41.09 & \bfseries 10.83 & \bfseries 51.24 \\
\midrule
\multirow{2}{*}{122B} & Before  & 12.47 & \bfseries 15.51 & \bfseries 64.48 & 42.35 & 12.15 & \bfseries 51.67 \\
 & After  & \bfseries 12.41 & 15.94 & 63.84 & \bfseries 39.64 & \bfseries 11.64 & 51.19 \\
\bottomrule
\end{tabular*}
\end{table}

\FloatBarrier

\subsection{Dependency Modeling Results}
\label{app:dependency_results}

\textbf{Setup}:
We evaluate LongCap-35B-A3B and LongCap-122B-A10B with event-dependency
supervision on LongCapBench-TAC-Short and LongCapBench-TAC-Long.
DeepSeek and Gemini independently annotate dependencies on the benchmark
reference events. These text-derived silver graphs are not audio-verified ground truth.
A separate DeepSeek matcher aligns local captions and timestamps with positive
temporal overlap, hiding IDs, ancestors, and relations; the same event
correspondence serves both references. Native events and arcs are retained without rewriting;
an arc must resolve to a unique earlier declaration of its typed ID.
Excluding predictions with negative timestamps leaves 1,492/1,495 short-form
clips for 35B/122B. Reference-event controls cover all 1,500 clips; every
condition includes all four long-form recordings in full. Empty predictions
remain in scoring.

\textbf{Metrics}:
\emph{Link support} (LS) averages the reference-supported fraction of each
emitted arc's endpoint coverage over all three relation types
(\texttt{same\_entity}, \texttt{continues}, and \texttt{recurs}).
Support may come from another earlier member of the same typed reference group;
invalid or ambiguous addresses and unknown relations receive zero.
\emph{Connectivity $F_1$} here measures direct and transitive
\texttt{same\_entity} agreement using interval-pair weights $w_{ij}=d_i d_j$,
where $d_i,d_j$ are interval durations. Pairs within the same original
component on both sides are excluded. With predicted and reference connected
masses $W_H,W_R$ and supported intersection $M$,
\begin{equation}
F_1=\frac{2M}{W_H+W_R}.
\label{eq:dependency_f1}
\end{equation}
Only resolved reference paths supply positive credit; unresolved paths remain
in the denominators, and missing matching judgments remain uncertainty.
Scores macro-average defined per-record values: LS is undefined without
emitted arcs; $F_1$ is undefined without positives on either side and zero
with positives on only one side. Eligible cohorts therefore differ.
Table~\ref{tab:dependency_overall} separates the references;
Fig.~\ref{fig:dependency} averages their two macro scores equally.

\begin{table}[!htbp]
\caption{Dependency agreement (\%, higher is better). Native denotes the
model's original links; +GLM denotes GLM-5.3-Flash links on the indicated
events. LS covers all relations; $F_1$ covers \texttt{same\_entity}.}
\label{tab:dependency_overall}
\centering
\small
\setlength{\tabcolsep}{4pt}
\renewcommand{\arraystretch}{1.08}
\sisetup{table-format=2.2, table-number-alignment=center}
\begin{tabular*}{\linewidth}{@{\extracolsep{\fill}}ll *{4}{S}@{}}
\toprule
\multirow{2}{*}{Benchmark} & \multirow{2}{*}{Condition}
& \multicolumn{2}{c}{LS} & \multicolumn{2}{c}{$F_1$} \\
\cmidrule(lr){3-4}\cmidrule(l){5-6}
& & {DeepSeek} & {Gemini} & {DeepSeek} & {Gemini} \\
\midrule
Short form & 35B native & 54.40 & 55.71 & 62.34 & 64.39 \\
 & 122B native & 58.54 & 59.95 & 67.20 & 68.98 \\
\addlinespace[2pt]
 & Reference + GLM & 70.27 & 72.29 & 69.60 & 71.26 \\
 & 35B + GLM & 57.38 & 58.83 & 66.16 & 68.06 \\
 & 122B + GLM & 58.16 & 59.56 & 66.43 & 68.12 \\
\midrule
Long form & 35B native & 9.30 & 9.73 & 4.99 & 3.39 \\
 & 122B native & 22.86 & 24.66 & 16.47 & 13.06 \\
\addlinespace[2pt]
 & Reference + GLM & 49.56 & 52.62 & 22.87 & 16.44 \\
 & 35B + GLM & 32.13 & 35.00 & 18.35 & 16.43 \\
 & 122B + GLM & 35.44 & 38.84 & 24.09 & 20.46 \\
\bottomrule
\end{tabular*}
\end{table}
\FloatBarrier

\textbf{Results and GLM controls}:
Averaging the two references, the models' original long-form $F_1$ increases from
4.2\% (35B) to 14.8\% (122B), while both remain well below their
short-form scores (63.4/68.1\%).
For controls, GLM-5.3-Flash labels benchmark reference events and each
model's predicted events using the same labeling procedure. For predictions,
we replace the original links while preserving captions, timestamps, duplicate
events, and the evaluated clips. Chronological sorting and unique working IDs also
remove ambiguity in event identifiers. Relabeling raises long-form $F_1$ to
17.4/22.3\% (35B/122B), combining new links with address normalization;
these gains do not isolate learned dependency semantics or a model-size effect.
GLM on reference events reaches 70.4\% on short-form and 19.7\% on long-form
audio, so low long-form agreement persists even with reference events.
This is an agreement baseline, not a performance ceiling. The four-recording
sample, incomplete matching, and ambiguous native IDs limit interpretation
as long-range memory accuracy.

\end{document}